\documentclass[conference]{IEEEtran}
\usepackage{times}
\IEEEoverridecommandlockouts                              

\usepackage[numbers]{natbib}
\usepackage{multicol}
\usepackage[bookmarks=true]{hyperref}

\usepackage{amsmath} 
\usepackage{amssymb}  
\usepackage{breqn}
\usepackage{bm}

\usepackage[utf8]{inputenc}
\usepackage{booktabs}
\usepackage{multirow}
\usepackage{subcaption} 
\usepackage[font=small,labelfont=bf]{caption} 
\usepackage{array}
\usepackage{xspace}
\usepackage{flushend}
\usepackage{algorithm}
\usepackage{algorithmic}

\usepackage{graphicx}
\usepackage{float}
\usepackage[dvipsnames]{xcolor}
\usepackage{caption}
\usepackage{subcaption}
\usepackage{bm}
\usepackage{amssymb} 
\usepackage{color}
\usepackage{wrapfig}
\usepackage{textcomp}
\newcommand{\shrinka}{\def\baselinestretch{0.993}\large\normalsize}

\title{\LARGE \bf
DefGraspSim: Simulation-based grasping of 3D deformable objects}

\author{Isabella Huang$^{1, 2}$, Yashraj Narang$^{2}$, Clemens Eppner$^{2}$, Balakumar Sundaralingam$^{2}$, Miles Macklin$^{2}$, \\ Tucker Hermans$^{2, 3}$, Dieter Fox$^{2, 4}$
\thanks{$^{1}$Department of Electrical Engineering and Computer Sciences, University of California, Berkeley, USA;$^{2}$NVIDIA Corporation, Seattle, USA;$^{3}$School of Computing, University of Utah, Salt Lake City, USA;$^{4}$Paul G. Allen School of Computer Science \& Engineering, University of Washington, Seattle, USA}}
\begin{document}
\shrinka
\maketitle
\shrinka
\thispagestyle{empty}
\pagestyle{empty}

\section{Introduction}
From clothing, to plastic bottles, to humans, deformable objects are omnipresent in our world. A large subset of these are \textit{3D deformable objects} (e.g., fruits, internal organs, and flexible containers), for which dimensions along all 3 spatial axes are of similar magnitude, and significant deformations can occur along any of them \cite{Sanchez2018IJRR}. Robotic grasping of 3D deformables is underexplored relative to rope and cloth, but is critical for applications like food handling \cite{Gemici2014IROS}, robotic surgery \cite{Smolen2009ICACHI}, and domestic tasks \cite{Sanchez2018IJRR}. Compared to rigid objects, grasping 3D deformable objects faces 4 major challenges.

First, classical analytical metrics for grasping rigid objects (e.g., force/form closure) do not typically consider deformation of the object during or after the grasp \cite{Sanchez2018IJRR}. Yet, deformations significantly impact the contact surface and object dynamics.

Second, existing grasp strategies for rigid objects may not directly transfer to 3D deformables, as compliance can augment or reduce the set of feasible grasps. For example, we may grasp a soft toy haphazardly; however, if the toy were rigid, it would no longer conform to our hands, and many grasps may become unstable. Conversely, we may grasp a rigid container haphazardly; however, if the container were flexible, grasps along its faces may crush its contents.

Third, the definition of a successful grasp on a 3D deformable is highly dependent on object properties, such as fragility and compliance. Thus, grasp outcomes must be quantified by diverse \textit{performance metrics}, such as stress, deformation, and stability. Performance metrics may also compete (e.g., a stable grasp may induce high deformation).

Fourth, performance metrics may be partially or fully unobservable (e.g., volumetric stress fields), requiring estimation in the real world. Previous works have typically formulated \textit{quality metrics}, which we refer to more generally as \textit{grasp features}: simple quantities that a robot can measure before pickup that can predict performance metrics. Whereas many grasp features have been proposed for rigid objects, analogous features for deformable objects are limited.

\begin{figure}
\centering
\includegraphics[scale=0.30, trim={0cm 0cm 0cm 0cm},clip]{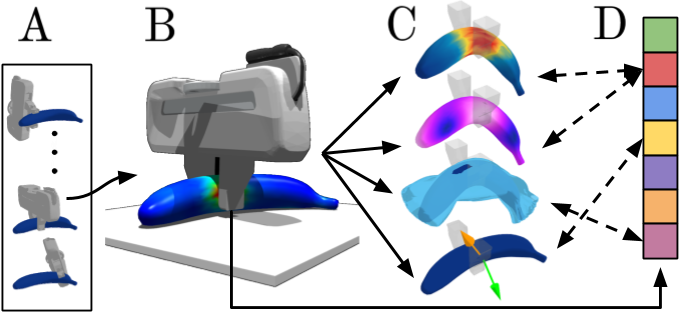}
\caption{(A) For a broad set of candidate grasps on a deformable object, (B) we simulate the object's response with FEM, (C) measure performance metrics (e.g., stress, deformation, controllability, instability), and (D) identify pre-pickup grasp features that are correlated with the metrics. Our simulated dataset contains 34 objects, 6800 grasp evaluations, and $1.1M$ unique measurements. 
}
\vspace{-16pt}
\label{fig:front_figure}
\end{figure}

Given these complexities, we conduct a large-scale simulation-based study of 3D deformable object grasping (Fig.~\ref{fig:front_figure}). Simulation affords multiple advantages: it extends analytical methods through accurate modeling of object deformation, enables safe execution of experiments, and provides full observability of performance metrics. For an overview of existing literature on deformable object modeling, grasp performance metrics, and grasp features, see \textbf{Appendix~\ref{app:related_works}}.

We then leverage the corotational finite element method (FEM) to conduct several thousand grasping simulations on 3D deformables varying in geometry and elasticity. First, we simulate grasping on 6 object primitives; for each primitive, we methodically describe the effects of different grasps on performance metrics, and quantify the ability of each feature to predict each metric. Given the small number of prior works on grasping 3D deformables, this examination of primitives establishes valuable physical intuition.

Furthermore, we provide our live dataset of 34 objects, 6800 grasp evaluations, and $1.1M$ corresponding measurements. We also release our codebase, which can automatically perform our exhaustive set of FEM-based grasp evaluations on 3D objects of the user's choice. Finally, we provide an interactive visualizer of our results and a video of our simulated and real-world experiments.\footnote{\url{https://sites.google.com/nvidia.com/defgraspsim}} We believe these contributions are an important conceptual and practical first step towards developing a complete learning and planning framework for grasping 3D deformables.



\vspace{-6pt}
\begin{figure}[h]
\centering
\frame{\includegraphics[scale=0.30]{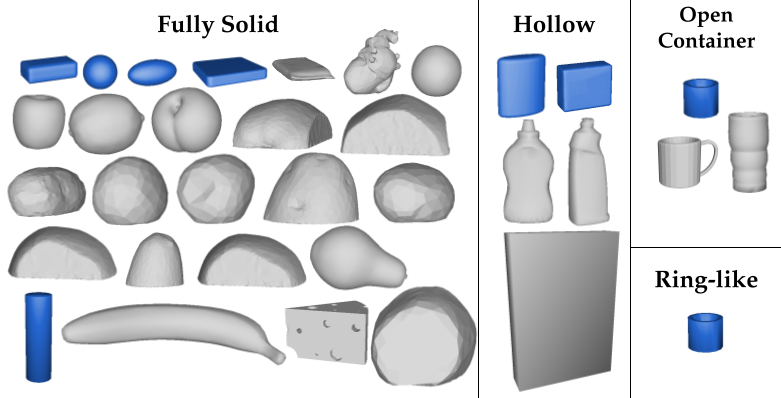}}
\caption{The 34 evaluated objects grouped by geometry and dimension (shown to scale). Objects in blue are self-designed primitives; those in gray are scaled models from open datasets \cite{Calli2015RAM,Wu2015CVPR,thingiverse, ybjDataset}.
}
\label{fig:object_categories}
\vspace{-18pt}
\end{figure}
\section{Grasp Simulator and Experiments} \label{sim}
We use Isaac Gym \cite{isaacgym} to simulate grasps of the widely-used Franka parallel-jaw gripper on 3D deformable objects. To generalize to other parallel-jaw grippers, proprietary gripper features are removed. The 3D deformables consist of 34 object primitives and real-world models, with geometry and dimensionality distributed across a broad categorization (Fig. \ref{fig:object_categories}), and elastic moduli distributed from $10^4$ to $10^9$~$Pa$ (i.e., human skin to hard plastic). 
The deformables are represented as tetrahedral meshes and simulated using 3D co-rotational FEM, with resulting equations solved using a GPU-based Newton method \cite{macklin2019}; a recent study has validated the simulator accuracy \cite{Narang2021Latent}. Addressing the limitations of classical analytical approaches, the simulator explicitly models complex object geometry, object deformation, gripper-object dynamics, and large perturbations. Simulations execute at $5$-$10$~$fps$, and in total, the dataset required 2080 GPU hours. Further details on the simulator itself are in \textbf{Appendix~\ref{appendix_sim}}.

Within Isaac Gym, we perform grasping experiments on 34 objects using a simulated Franka parallel-jaw gripper. 
For each object, a diverse set of 50 candidate grasps is generated using an antipodal sampler~\cite{EppnerISRR2019}.
For each grasp, 4 tests are executed: \textit{pickup, reorientation, linear acceleration, and angular acceleration}. Details of each test are in \textbf{Appendix~\ref{appendix_experiment}}. 
\section{Grasp Performance Metrics and Features}

\label{sec:metrics}
During the preceeding experiments, we measure 7 performance metrics to evaluate grasp outcomes, and 7 grasp features that correlate with the metrics. Details and motivation for these quantities are in \textbf{Appendix \ref{app:performance_metrics}}.

\vspace{0.2cm}

\begin{minipage}[t]{0.39\linewidth}   
Metrics:

\begin{itemize}
\item Pickup success
\item Stress
\item Deformation
\item Strain energy
\item Linear instability
\item Angular instability
\item Def. controllability
\end{itemize}
  \end{minipage}
\begin{minipage}[t]{0.55\linewidth}
Features:

\begin{itemize}
\item Contact patch distance to centroid 
\item Contact patch perpendicular distance to centroid
\item Number of contact points
\item Contact patch distance to finger edge
\item Gripper squeezing distance
\item Gripper separation 
\item Alignment with gravity
\end{itemize}
  \end{minipage}

\vspace{0.2cm}

\begin{figure}
     \centering
     \begin{subfigure}[b]{.32\columnwidth}
         \centering
         \includegraphics[scale=0.18, trim={0cm 0cm 44cm 0cm},clip]{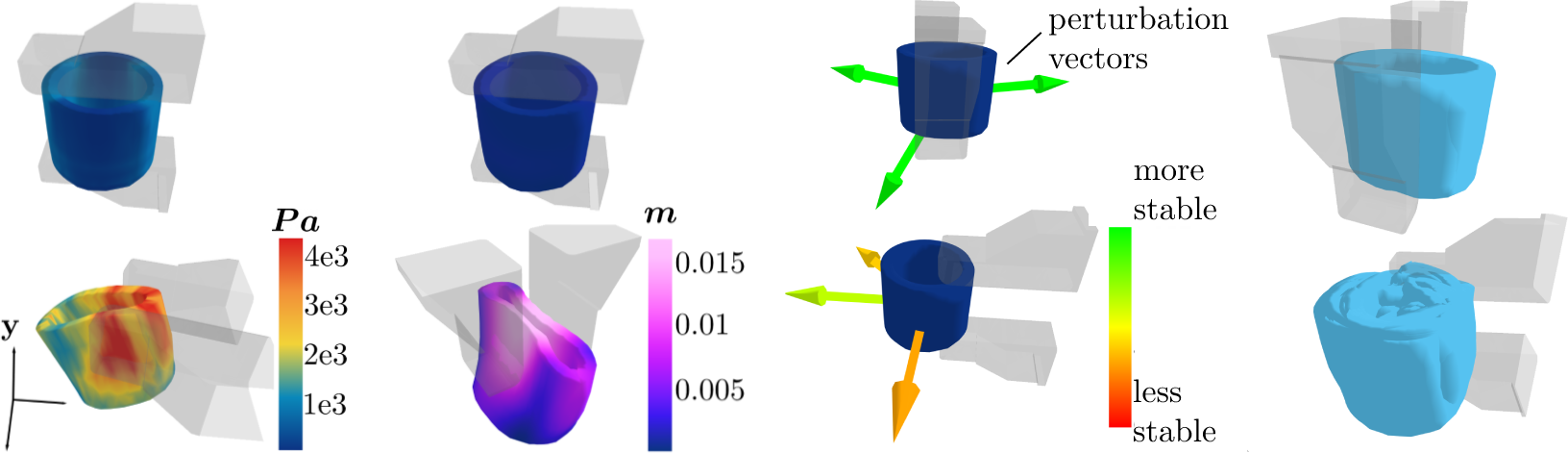}
         \caption{von Mises stresses}
         \label{fig:cup_short_a}
     \end{subfigure}
     \hfill
     \begin{subfigure}[b]{.32\columnwidth}
         \centering
         \includegraphics[scale=0.18, trim={14cm 0cm 30cm 0cm},clip]{figures/teasers/cup_shortened_3.png}
         \caption{Deformation norms}
         \label{fig:cup_short_b}
     \end{subfigure}
     \hfill
     \begin{subfigure}[b]{.32\columnwidth}
         \centering
         \includegraphics[scale=0.18, trim={29cm 0cm 12cm 0cm},clip]{figures/teasers/cup_shortened_3.png}
         \caption{Angular instability}
         \label{fig:cup_short_c}
     \end{subfigure}
     \caption{Low (top) and high-valued (bottom) grasps for various metrics on a cup.}
     \label{fig:cup_shortened}
\vspace{-16pt}
\end{figure}

In \textbf{Appendix~\ref{primitives}}, we show grasping results for 6 of 9 object primitives (Fig.~\ref{fig:object_categories}) over a wide range of elastic moduli, and use physical reasoning to interpret the relationships between metrics and features. Example results of various metrics on a cup are visualized in Fig.~\ref{fig:cup_shortened}. An analysis of grasp feature importance in predicting these metrics is also performed. 

\section{Sim-to-Real Accuracy}
We compare simulated and real-world grasp responses, and demonstrate that grasps performed on simulated blocks of tofu and latex tubing induce highly analogous responses on their real-world counterparts. All material parameters in simulation are acquired from reported values of common materials, and \textit{are not tuned to match the real outcomes}. Example correspondences are shown in Fig.~\ref{fig:hollow_tube_grasps_shortened}, with details in \textbf{Appendix~\ref{app:sim2real}}.


\begin{figure}[t]
\centering
\includegraphics[width=0.845\linewidth, trim={0cm 0cm 0cm 0cm},clip]{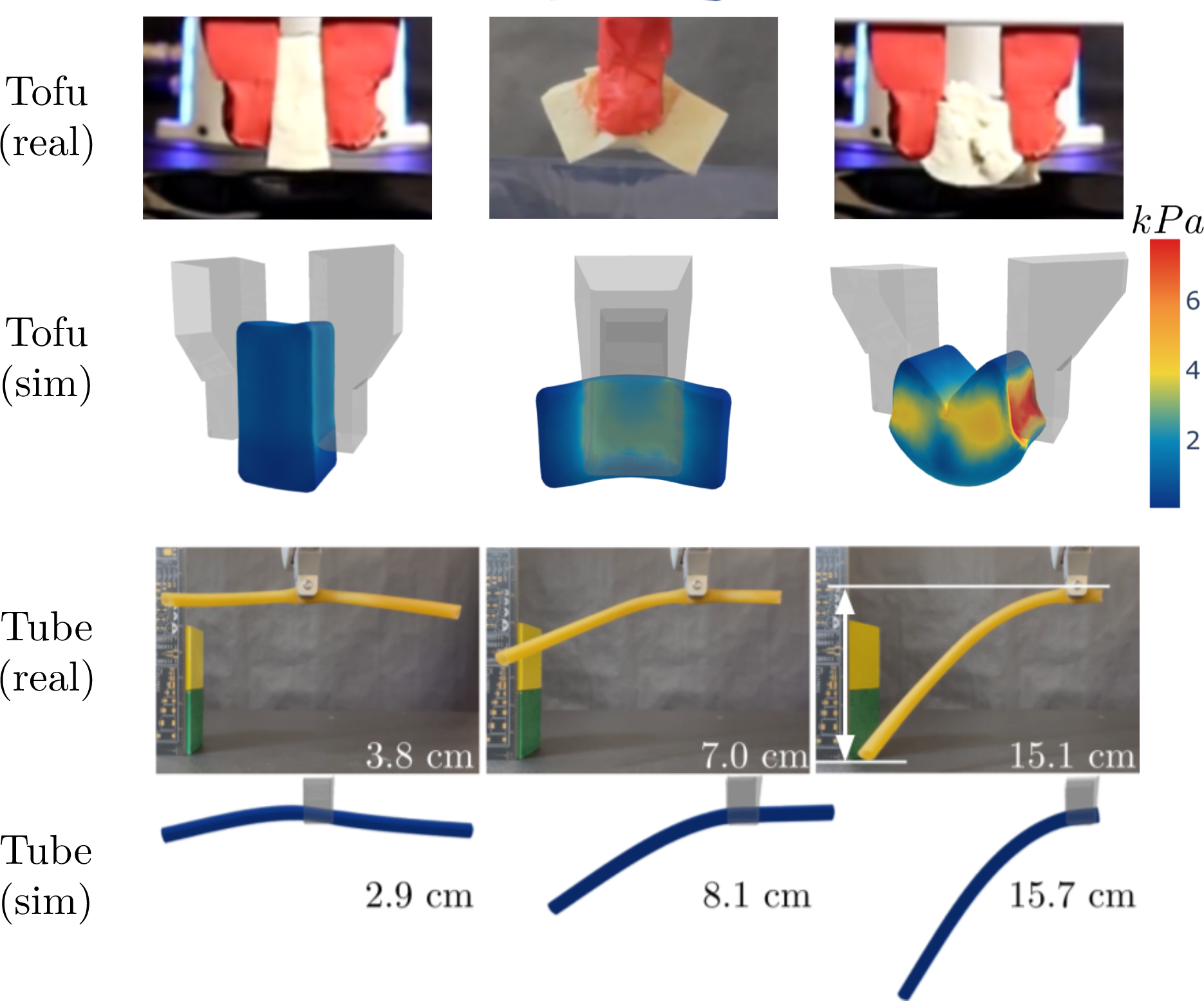}
\caption{Examples of 3 different grasps on a real and simulated block of tofu (top 2 rows), and 3 different grasps on a real and simulated latex tube (bottom 2 rows).}
\label{fig:hollow_tube_grasps_shortened}
\end{figure}

\section{Conclusions and Future Applications}
We conduct a battery of grasp simulations on 3D deformable objects and analyze grasp outcomes across performance metrics.
We propose grasp features and demonstrate their ability to predict the performance metrics. Our physical experiments also validate the accuracy of our large-scale simulations. We release our dataset of $1.1M$ measurements for further study, along with software that executes our experiments on arbitrary objects and material parameters. Based on our analysis of object primitives, we also identify several fundamental physical trends. We direct the reader to Appendix~\ref{primitives} for these findings. For future work, DefGraspSim can be used to:

\begin{itemize}
    \item Study new, high-dimensional features and metrics (e.g., geometric encoding of object shape and contact patch, full stress distributions, etc.), since raw nodal data is accessible within DefGraspSim
    \item Scale up grasping experiments to create task-oriented planners (e.g., to minimize deformation during transport)
    \item Perform rigorous, direct comparisons between simulation and reality on custom objects of interest
    \item Extend vision-based measurements from RGB-D images to interface with real sensors
    \item Complement real-world system identification (e.g. tactile probing or visual shape completion) 
    \item Improve grasp planning robustness to uncertainty in object properties (e.g., via domain randomization)
\end{itemize}

\clearpage
\newpage
\bibliographystyle{plainnat}
\bibliography{references}  
\newpage

\begin{appendices}

\section{Related Work}\label{app:related_works}
\noindent\textbf{Modeling techniques}. With over three decades of development, methods in rigid-object grasp planning range from model-based approaches using exact geometries \cite{Sastry1988IJRA,Ferrari1992ICRA,Miller2003ICRA} to data-driven approaches without full models \cite{Lenz2013IJRR,Dang2012a,Mahler2017CORR,Kopicki2016,mousavian2019,lu-ram2020-grasp-inference}. Rigid-body grasping simulators such as GraspIt!~\cite{Miller2004RAM} and OpenGRASP~\cite{Len2010OpenGRASP} have been used to develop many of these algorithms. Libraries like Bullet~\cite{coumans2019} and MuJoCo~\cite{Todorov2012IROS} can also model deformable ropes and cloths using rigid-body networks with compliant constraints. Such simulators have enabled real-world success in rope tying \cite{Lee2014IROS}, string insertion \cite{Wang2015ICRA}, cloth folding \cite{Maitin-Shepard2010ICRA,Li2015IROS}, and dressing \cite{Koganti2015IROS,Clegg2020RAL}. For 3D deformable objects, rigid-body approximations can lead to efficient simulations \cite{pozziefficient}; however, continuum models are preferred, as they can represent large deformations and allow consistent material parameters without an explicit model-fitting stage \cite{duriez.13}. 3D continuum models include Kelvin-Voigt elements governed by nonlinear PDEs~\cite{Howard2000AR}, mass-spring models~\cite{Lazher2014ICR}, 2D FEM for planar and ring-like objects~\cite{Jia2014IJRR}, and gold-standard 3D FEM~\cite{Lin2015IJRR}. However, many powerful FEM simulators used in engineering and graphics (e.g., Vega~\cite{Vega}) do not feature infrastructure for robotic control, such as built-in joint control.
For comprehensive reviews of 3D deformable modeling techniques, please refer to \cite{Arriola2020FRAI, YinScienceRobotics2021}.

\noindent\textbf{Performance metrics}. Prior works have evaluated 3D deformable-object grasps using performance metrics based on pickup success, strain energy, deformation, and stress. 
Success-based metrics include the minimal squeezing force required by a particular grasp, which can be calculated via real-world iterative search~\cite{Howard2000AR} and FEM~\cite{Lazher2014ICR,Lin2015IJRR}. Success is dependent on both object geometry and stiffness (e.g., a cone can be picked up only when soft enough to deform to the gripper) \cite{Lin2015IJRR}. Metrics based on strain energy (i.e., elastic potential energy stored in the object) have served as proxies for an object's stability against external wrenches. In 2D the \textit{deform} closure metric generalizes rigid form closure \cite{Bicchi1995IJRR} and quantifies the positive work required to release an object from a grasp \cite{Goldberg2005IJRR}. It is optimized by maximizing strain energy without inducing plastic deformation. Similarly, for thin and planar 2.5D objects, grasps have been selected to maximize strain energy under a fixed squeezing distance~\cite{Jia2014IJRR}. Deformation-based metrics have also been proposed for cups and bottles to detect whether contents are dislodged during lifting and rotation \cite{Xu2020ICRA}. Finally, stress-based metrics have been proposed to avoid material fracture, but were evaluated only on rigid objects \cite{Pan2020ICRA}. 

\noindent\textbf{Grasp features}. Many grasp features to predict grasp performance have been previously investigated on rigid objects. Features include force and form closure \cite{Ferrari1992ICRA} and grasp polygon area \cite{Mirtich1994ICRA}, and their predictive accuracy has been tested under different classification models \cite{Rubert2017IROS}. A thorough survey in rigid grasping features can be found in \cite{Roa2014AR}. However, grasp features for deformable objects have only been explored in one study, which measured the work performed on containers during grasping to predict whether its liquid contents would be displaced \cite{Xu2020ICRA}.

\section{Finite Element Simulation Details}\label{appendix_sim}

FEM is a variational numerical technique that divides complex geometrical domains into simple subregions and solves the weak form of the governing partial differential equations over each region.
In FEM simulation, a deformable object is represented by a volumetric mesh of \textit{elements}; the object's configuration is described by the element vertices, known as \textit{nodes}. We use Isaac Gym's\cite{isaacgym} co-rotational linear constitutive model of the object's internal dynamics coupled to a rigid-body representation of the robotic gripper via an isotropic Coulomb contact model~\cite{stewart2000rigid}. A GPU-based Newton method performs implicit integration by solving a nonlinear complementarity problem~\cite{macklin2019}. Unlike classical analytical models, this technique explicitly models complex object and gripper geometry, deformation and dynamics, as well as large kinematic and kinetic perturbations.
At each timestep, the simulator returns element stress tensors and nodal positions, which are used to calculate grasp metrics. With sufficiently small timesteps and high mesh density, FEM predictions for deformable solids can be extremely accurate \cite{Reddy2019Book, Narang2020RSS}. We simulate at the high frequency of $1500 Hz$ resulting in experiments running at 5-10 frames per second. In total, the dataset for this paper required approximately 1400 GPU hours to generate.

We evaluate a set of 23 3D deformable objects comprising both simple object primitives and complex real-world models, categorized by geometry and dimension (Fig. \ref{fig:object_categories}). We process object surface meshes in Blender to smooth sharp edges and reduce node count, then convert them into tetrahedral meshes using fTetWild \cite{ftw}. All objects have homogeneous material properties due to current limitations of Isaac Gym. 
Objects have density $\rho = 1000 \frac{kg}{m^3}$, Poisson's ratio $\nu = 0.3$, coefficient of friction $\mu = 0.7$, and Young's modulus $E \in \mathcal{E} = \{2e4, 2e5, 2e6, 2e9\} Pa$. $\mathcal{E}$ covers a wide range of real materials, from human skin ($\sim$$10^4 Pa$) to ABS plastic ($\sim$$10^9 Pa$). Values below $\sim$$10^3 Pa$ are excluded due to increased interpenetration effects, as are values above $\sim$$10^9 Pa$ due to no substantial differences arising.
The desired squeezing force on an object is $F_{p} = 1.3 \times \frac{mg}{\mu}$ (where $m$ is mass and $g$ is gravity), the frictional force required to support the object's weight with a factor of safety. For a fixed $E$, increasing $\mu$ decreases $F_p$ as well as the induced deformation. This effect is essentially the same as if $\mu$ is fixed while $E$ is increased, since an elastically stiffer object will also deform less for the same $F_p$ applied. Thus, we fix $\mu$ and vary $E$.

\section{Performance Metric Details}
\label{app:performance_metrics}
\noindent\textbf{Pickup success:} A binary metric measuring whether an object is lifted from a support plane. 

\noindent\textbf{Stress:} The element-wise stress field of an object when picked up. Exceeding material thresholds (e.g., yield stress, ultimate stress) leads to permanent deformation, damage, or fracture; examples include creasing of boxes, bruising of fruit, 
and perforation of organs. 

We convert each element's stress tensor into von Mises stress, a scalar quantity that quantifies whether an element has exceeded its yield threshold. We then measure the maximum stress over all elements, since real-world applications typically aim to avoid damage at any point.

\noindent\textbf{Deformation:} The node-wise displacement field of the object from pre- to post-pickup, neglecting rigid-body transformations. Deformation must often be minimized (e.g., on flexible containers with contents that can be damaged or dislodged). To compute this field, the difference between the pre- and post-pickup nodal positions is calculated, the closest rigid transform is determined \cite{hornung2017}, and the transform is subtracted. We compute the $\ell^2$ norm of each node's displacement and measure the maximum value over all nodes.

\noindent\textbf{Strain energy:} The elastic potential energy stored in the object (analogous to a Hookean spring). Conveniently, this metric penalizes both stress and deformation. The strain energy is given by $U_e = \int_V \sigma^T \epsilon dV$, where $\sigma$, $\epsilon$, and $V$ are the stress tensor, strain tensor, and volume, respectively.

\noindent\textbf{Linear and angular instability:} We define instability as the minimum acceleration applied to the gripper (\textit{along} or \textit{about} a vector for linear and angular instability, respectively) at which the object loses contact (i.e., separates along the gripper normal, or slides out of the gripper). This measures how easily an object is displaced from the grasp under external forces.

\begin{figure}
\centering
\includegraphics[scale=0.09]{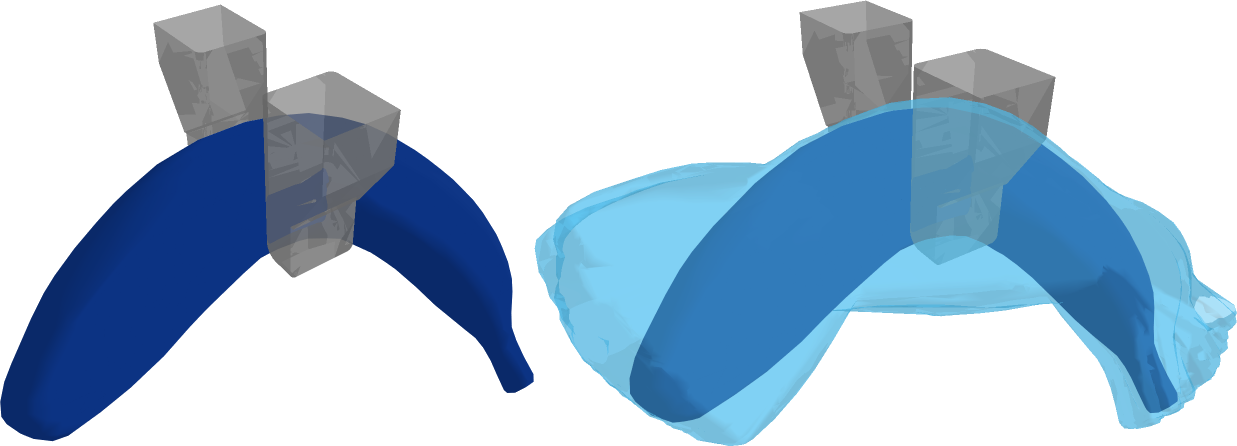}
\caption{Illustration of deformation controllability. (Left) A banana under pickup; (Right) The union of all banana configurations achieved under reorientation, superimposed in light blue. 
}
\label{fig:def_banana}
\vspace{-18pt}
\end{figure}

\noindent\textbf{Deformation controllability:} We define deformation controllability as the maximum deformation when the object is reoriented under gravity (e.g., an illustration of an example object's shape changes induced during reorientation under gravity is pictured in Fig.~\ref{fig:def_banana}). Depending on the task, it may be useful to either minimize or maximize deformation controllability. For example, to reduce the effects of post-grasp reorientation on deformation, minimizing this metric allows the object to behave rigidly after pickup. Alternatively, to augment the effects of post-grasp reorientation (e.g., during insertion of endoscopes), we may maximize it instead. 
Our notion of deformation controllability is different from the classical notion (i.e., the ability to achieve any robot state in finite time). Here, we are not modifying robot controllability by changing actuation, but modifying object controllability by changing the number of possible deformation states. 

\section{Grasp Feature Details}
\label{app:grasp_features}

The 7 grasp features are recorded after applying the grasping force $F_p$, but before pickup. They can all be measured by common sensors (e.g., encoders, cameras, and tactile arrays). They are summarized in Table~\ref{tab:1}, along with examples of references to existing works from which they are derived. Please refer to \cite{Roa2014AR} for a full review of grasp features on rigid objects.


\begin{table*}[t]
  \centering
  \begin{tabular}{|p{4cm}|p{1.5cm}|p{7cm}|p{2cm}|}
    \hline
    Feature & Abbreviation & What it quantifies & Usage in existing literature \\
    \hline 
    Contact patch distance to centroid 
    & \textit{pure\_dist} 
    & Distance from the center of each finger's contact patch to the object's center of mass (COM) (Fig. \ref{fig:qm}\subref{fig:qm1}), averaged over the two fingers.
    & \cite{Rubert2017IROS, DingITRA2001} \\
    \hline 
    Contact patch perpendicular distance to centroid 
    & \textit{perp\_dist} 
    & Perpendicular distance from the center of each finger's contact patch to the object's COM (Fig. \ref{fig:qm}\subref{fig:qm1}), averaged over the two fingers; quantifies distance from lines of action of squeezing force.
    & \cite{BalasubramanianICRA2010} \\
    \hline 
    Number of contact points 
    & \textit{num\_contacts} 
    & Number of contact points on each finger, averaged over the fingers; quantifies amount of contact made.
    & \cite{Rubert2017IROS, DingITRA2001} \\
    \hline 
    Contact patch distance to finger edge 
    & \textit{edge\_dist} 
    & Distance from each finger's distal edge to the center of its contact patch (Fig. \ref{fig:qm}\subref{fig:qm2}), averaged over the two fingers.
    & \cite{Feix2016ITHMS} \\
    \hline 
    Gripper squeezing distance
    & \textit{squeeze\_dist} 
    & Change in finger separation from initial contact to the point at which $F_p$ is achieved; quantifies local deformation applied to the object.
    & \cite{Xu2020ICRA} \\
    \hline 
    Gripper separation 
    & \textit{gripper\_sep} 
    & Finger separation upon achieving $F_p$; quantifies the thickness of material between the fingers at grasp.
    & \cite{Rubert2017IROS} \\
    \hline 
    Alignment with gravity 
    & \textit{grav\_align} 
    & Angle between the finger normal and the global vertical; grounds the grasp pose to a fixed frame (Fig. \ref{fig:qm}\subref{fig:qm2}).
    & \cite{Vina2016ICRA} \\
    \hline
  \end{tabular}
  \caption{Grasp features, their descriptions, and existing works from which they are derived.}
  \label{tab:1}
\end{table*}

\begin{figure}[t]
     \centering
     \begin{subfigure}[b]{.31\columnwidth}
         \centering
         \includegraphics[scale=0.2, trim={0cm 4cm 15cm 5.5cm},clip]{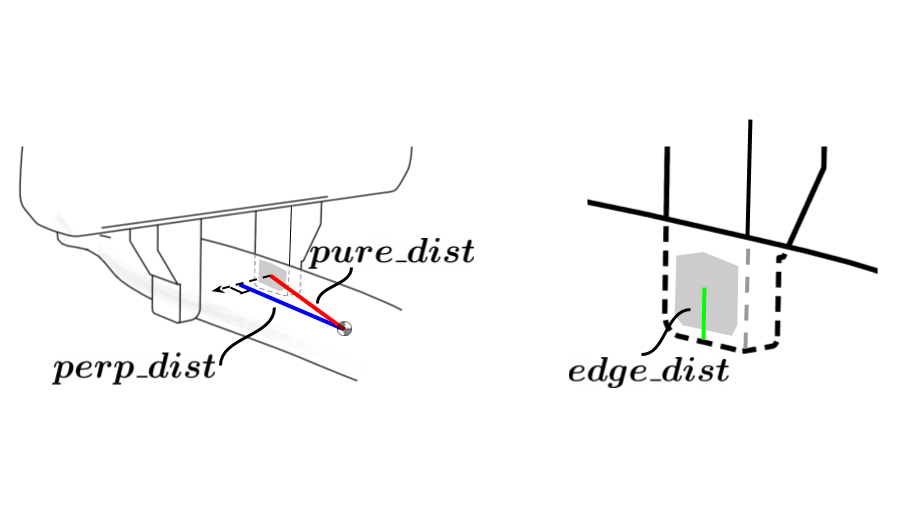}
         \caption{}
         \label{fig:qm1}
     \end{subfigure}
     \hfill
     \begin{subfigure}[b]{.31\columnwidth}
         \centering
         \includegraphics[scale=0.2, trim={20cm 4cm 3.5cm 5cm},clip]{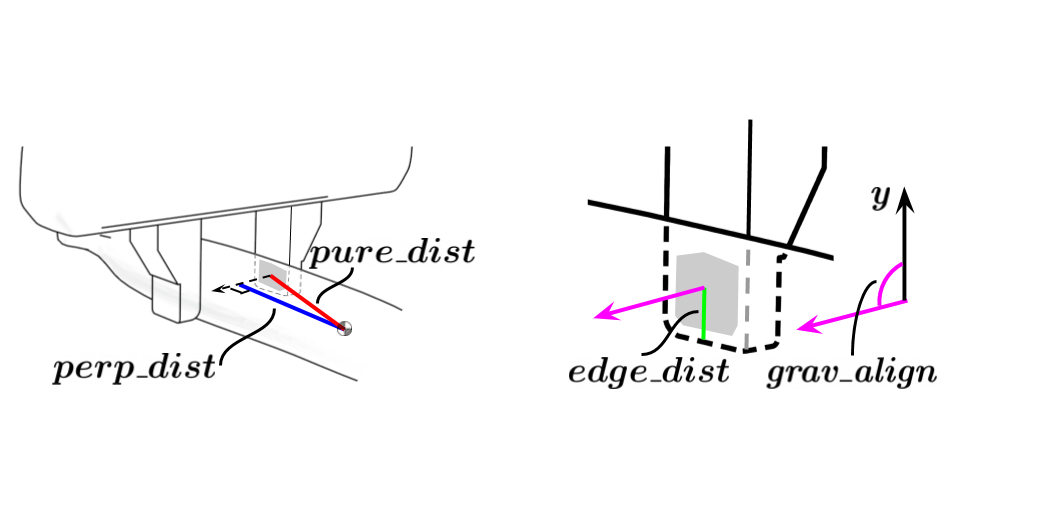}
         \caption{}
         \label{fig:qm2}
     \end{subfigure}
     \hfill
     \begin{subfigure}[b]{.31\columnwidth}
         \centering
         \includegraphics[scale=0.1, trim={2.5cm 1.5cm 4.5cm 1.5cm},clip]{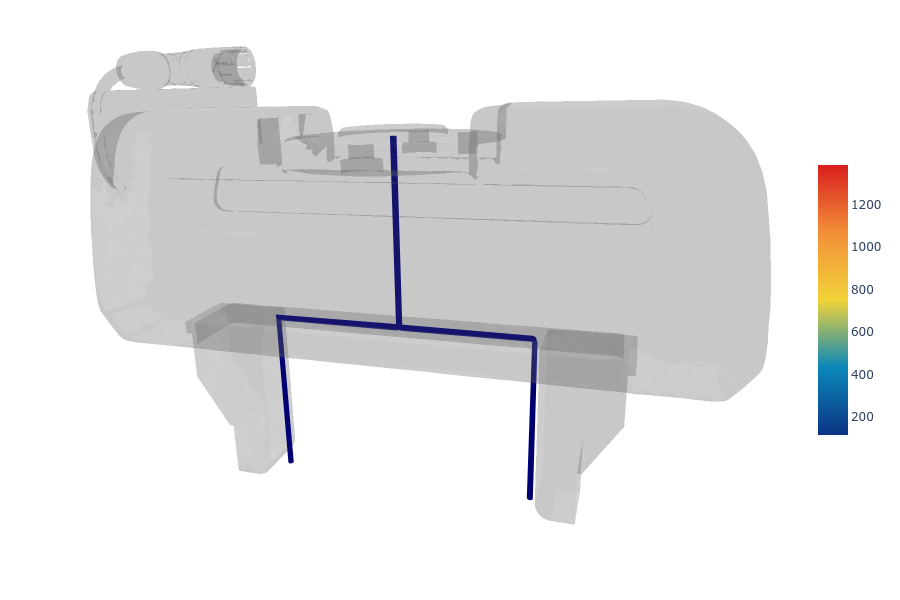}
         \caption{}
         \label{fig:skeleton}
     \end{subfigure}
        \caption{(ab) Four grasp features illustrated on a Franka gripper; (c) Line representation of gripper pose, used in later sections. 
        }
        \label{fig:qm}
\vspace{-18pt}
\end{figure}

\section{Experimental Simulation Details}\label{appendix_experiment}

\begin{figure*}[h]
\centering
\includegraphics[scale=0.4]{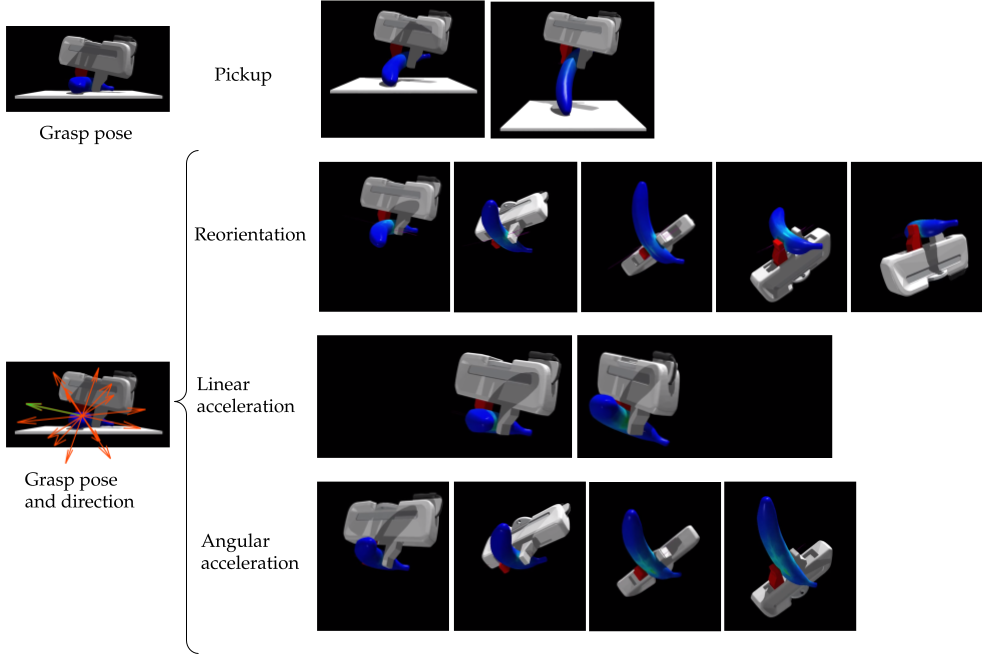}

\caption{Example frames from the execution of four different tests per grasp on a banana: pickup, reorient, twist (angular acceleration), and shake (linear acceleration).} 
\label{fig:grasp_tests}
\end{figure*}

For completeness and the ability to replicate our work we now explain additional details and settings used in our simulation-based experiments. We visualize the tests in Figure~\ref{fig:grasp_tests}.

Each object initially rests atop a horizontal plane; gripper collisions with the plane are disabled, as we want to test the full spatial distribution of grasps by allowing grasps to come from underneath regardless of collisions that would occur in the real world.

Prior to grasping, the pre-contact nodal positions and element stresses of the object are recorded. The gripper is then initialized at a candidate grasp pose. The gripper squeezes using a force-based torque controller to achieve $F_p$, with a low-pass filter applied to contact forces to mitigate numerical fluctuations. Once $F_p$ converges, the grasp features are measured. Then, one of the tests listed in Section~\ref{sec:sim-experiments} is executed depending on the performance metric to be evaluated.
 

\noindent\textbf{Pickup}. The platform is lowered to apply incremental gravitational loading to the object. Pickup is a success if the gripper maintains contact with the object for 5 seconds. If so, stress and deformation fields are recorded, and stress, deformation, and strain energy performance metrics are computed.

\noindent\textbf{Reorientation}. The gripper squeezing force is increased from $F_{p}$ to $F_{slip}$, the minimum force required to counteract rotational slip. The platform is lowered until the object is picked up. The gripper rotates the object to 64 unique reorientation states.
Stress and deformation fields are recorded at each state, and deformation controllability is computed as the maximum deformation over all states. $F_{slip}$ is estimated by approximating each gripper contact patch as 2 point-contacts that oppose the gravitational moment. The gripper rotates the object about each of 16 vectors regularly spaced in a unit 2-sphere at angles $k\pi/4, k \in [1..4]$ for a total of 64 unique reorientation states.

\noindent\textbf{Linear acceleration}.
The gripper linearly accelerates along the 16 unique direction vectors as in the reorientation test. Each vector has a complement pointing in the opposite direction; thus, this method generalizes the cyclic shaking tests from previous works~\cite{EppnerISRR2019}. The acceleration is recorded at which at least one finger loses contact with the object. Linear instability is computed as the average loss-of-contact acceleration over all directions. The robot moves at at $1000 \frac{m}{s}^3$ jerk in a gravity-free environment, corresponding to a linearly increasing acceleration. We impose a realistic upper acceleration limit of $50 \frac{m}{s}^2$ ($\approx 5g$).

\noindent\textbf{Angular acceleration}. The gripper now rotationally accelerates about 16 unique axes. Angular instability is computed as the average loss-of-contact acceleration over all axes. The robot accelerates at $2500 \frac{rad}{s}^3$ jerk; to mitigate undesired linear acceleration, the midpoint between the fingers is set as the center of rotation. The angular loss-of-contact threshold is limited at $1000 \frac{rad}{s}^2$ (i.e., the linear acceleration limit, scaled by the $0.04 m$ maximum finger displacement, which approximates the moment arm).

\noindent\textbf{Controller Details.} A contact force-based torque controller is used to achieve the desired squeezing forces during grasping. A low-pass filter is first applied to the contact force signals due to high frequency noise that prevails from small numerical fluctuations in position, especially at higher moduli. For the three tests involving post-pickup manipulation, the finger joints are frozen immediately after pickup to maintain the gripper separation.

\section{Grasp Performance on Primitives} \label{primitives}
This section presents detailed grasping results for 6 of the 9 object primitives (Fig.~\ref{fig:object_categories}) over a wide range of elastic moduli, and uses physical reasoning to interpret the relationships between metrics and features. Each primitive abstracts a large set of real-world objects.
The \emph{rectangular prism} primitive represents objects such as sponges, tofu, and rubber blocks. The \emph{spheroid} primitive is 
an ellipse revolved about its major axis; it represents objects such as rounded fruits and rubber balls. 
The \emph{cup} and \emph{ring} primitives have the same geometry, but the ring lacks a base. Cups have direct real-world analogues; rings represent objects such as flexible tubing and gaskets. The geometric stiffness profile of a cup varies with height, whereas that of a ring does not; however, both objects behave differently between side grasps and top-bottom grasps. The \emph{hollow flask} primitive has an ellipsoidal cross section and represents objects such as boxes and bottles. The \emph{cylinder} primitive represents objects such as bananas and rubber tubes.

We divide our analysis into two sections. We first showcase the types of grasps that produce low and high values for the performance metrics defined in Sec.~\ref{sec:metrics}. We then evaluate the ability of the grasp features defined in Sec.~\ref{sec:features} to predict these metrics. 
Additional visualizations are provided on our website.
\subsection{Performance Metric Results}

\begin{figure}[t]
     \centering
     \begin{subfigure}[b]{.32\columnwidth}
         \centering
         \includegraphics[scale=0.09, trim={0cm 0cm 60cm 0cm},clip]{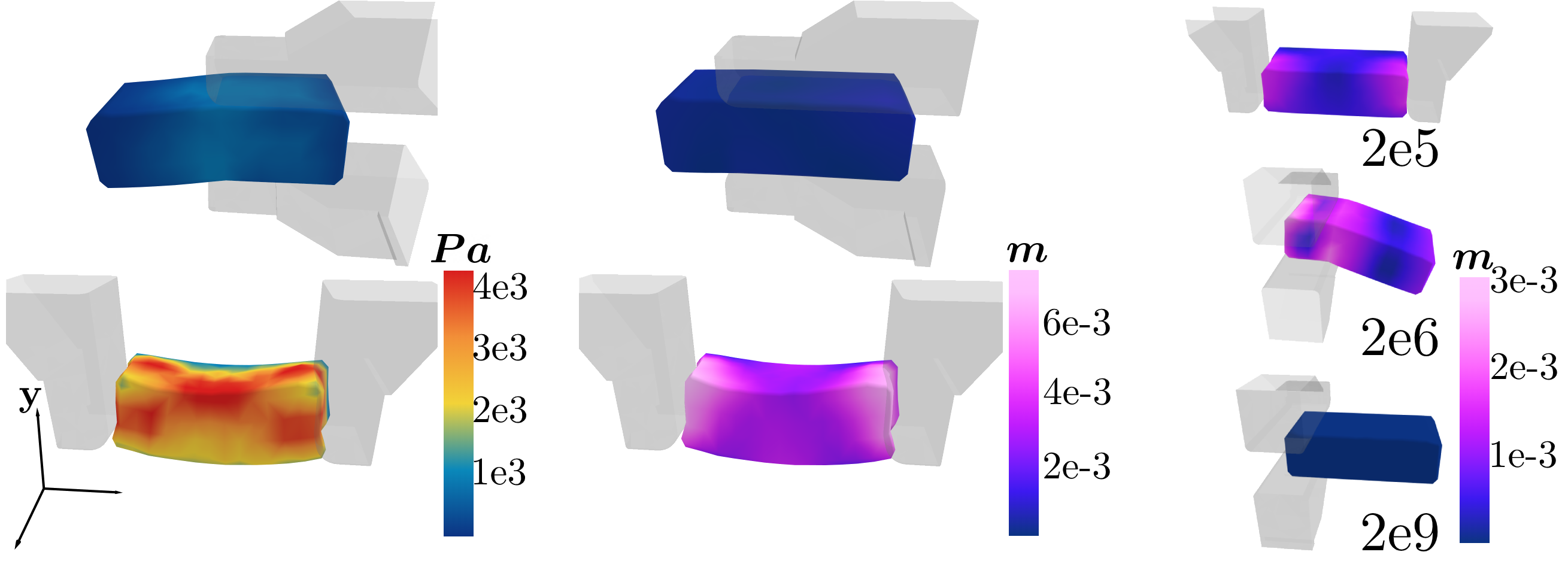}
         \caption{von Mises stresses}
         \label{fig:rect_a}
     \end{subfigure}
     \hfill
     \begin{subfigure}[b]{.32\columnwidth}
         \centering
         \includegraphics[scale=0.09, trim={34cm 0cm 26cm 0cm},clip]{figures/rectangle/rect_sd_big.png}
         \caption{Def. field norms}
         \label{fig:rect_b}
     \end{subfigure}
     \hfill
     \begin{subfigure}[b]{.32\columnwidth}
         \centering
         \includegraphics[scale=0.09, trim={69cm 0cm 0cm 0cm},clip]{figures/rectangle/rect_sd_big.png}
         \caption{High def. grasps}
         \label{fig:rect_c}
     \end{subfigure}
        \caption{(ab) Low (top) and high-valued (bottom) prism grasps for max stress and max deformation ($E{=}2e4$); (c) High-valued grasps for max deformation for additional elastic moduli.
        }
        \label{fig:rect_sd}
        \vspace{-18pt}
\end{figure}

\begin{figure}
     \centering
     \begin{subfigure}[b]{.33\columnwidth}
         \centering
         \includegraphics[scale=0.09, trim={0cm 0cm 61.5cm 0cm},clip]{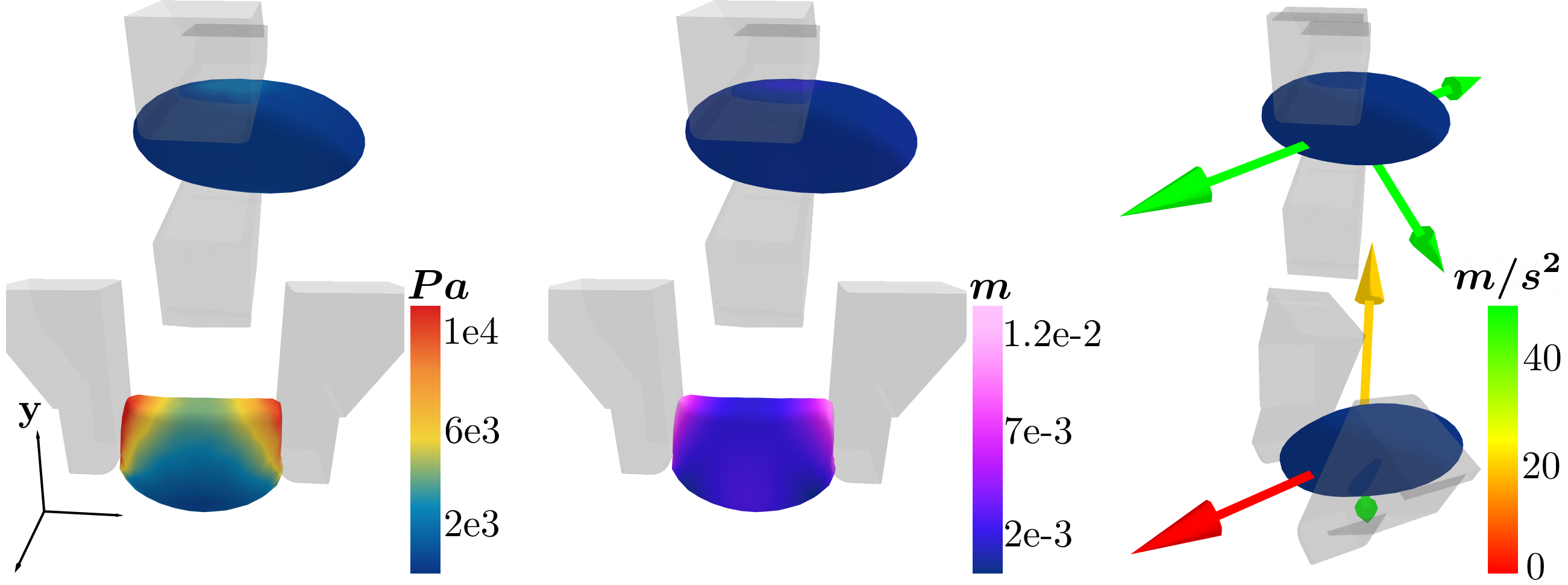}
         \caption{Max stress.}
         \label{fig:spheroid_a}
     \end{subfigure}
     \hfill
     \begin{subfigure}[b]{.33\columnwidth}
         \centering
         \includegraphics[scale=0.09, trim={32cm 0cm 26.5cm 0cm},clip]{figures/ellipsoid/ellipsoid_grasps.png}
         \caption{Max def.}
         \label{fig:spheroid_b}
     \end{subfigure}
     \hfill
     \begin{subfigure}[b]{.30\columnwidth}
         \centering
         \includegraphics[scale=0.09, trim={65cm 0cm 0cm 0cm},clip]{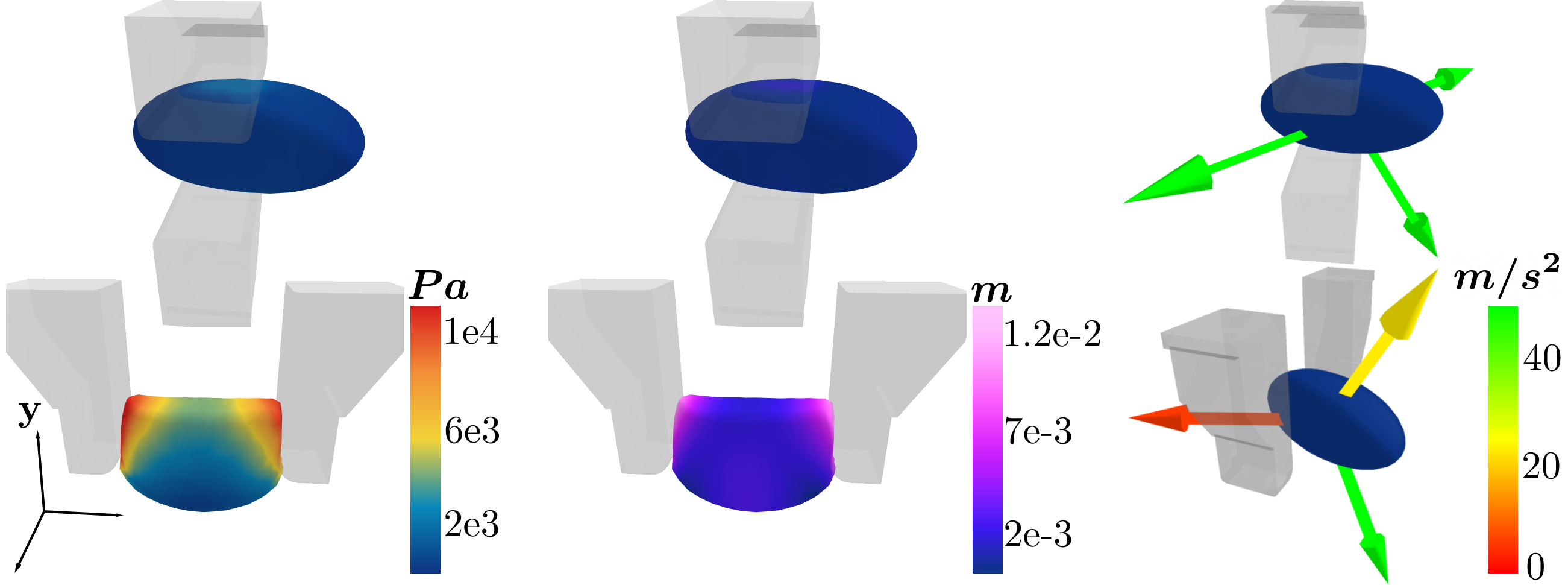}
         \caption{Lin. instability.}
         \label{fig:spheroid_c}
     \end{subfigure}
        \caption{Low (top) and high-valued (bottom) spheroid grasps ($E{=}2e4$).}
        \label{fig:spheroid_grasps}
\end{figure}

\noindent\textbf{Stress and Deformation Metrics.} We first examine the stress and deformation responses, which follow similar (but not identical) trends.
For the prism and spheroid, grasps with low values for maximum stress and deformation consistently squeeze along the shortest dimension  (Fig.~\ref{fig:rect_sd} and Fig.~\ref{fig:spheroid_grasps}ab). In this configuration, the gripper achieves the highest contact area, lowering compressive stress. Furthermore, the smallest amount of material is between the gripper fingers; analogous to springs in series, a smaller material thickness has a higher equivalent stiffness, reducing deformation. For the prism, grasps with the highest values for maximum stress consistently squeeze along the long axis. The object tends to buckle, increasing bending stress. However, grasps with the highest values for maximum deformation vary with $E$ (Fig.~\ref{fig:rect_sd}c). The highest-valued grasps at $E{=}\{2e6, 2e9\}$ squeeze the end of the prism, allowing cantilever bending under gravity.

\begin{figure}
     \centering
     \begin{subfigure}[b]{.38\columnwidth}
         \centering
         \includegraphics[scale=0.085, trim={0cm 1.5cm 68cm 0cm},clip]{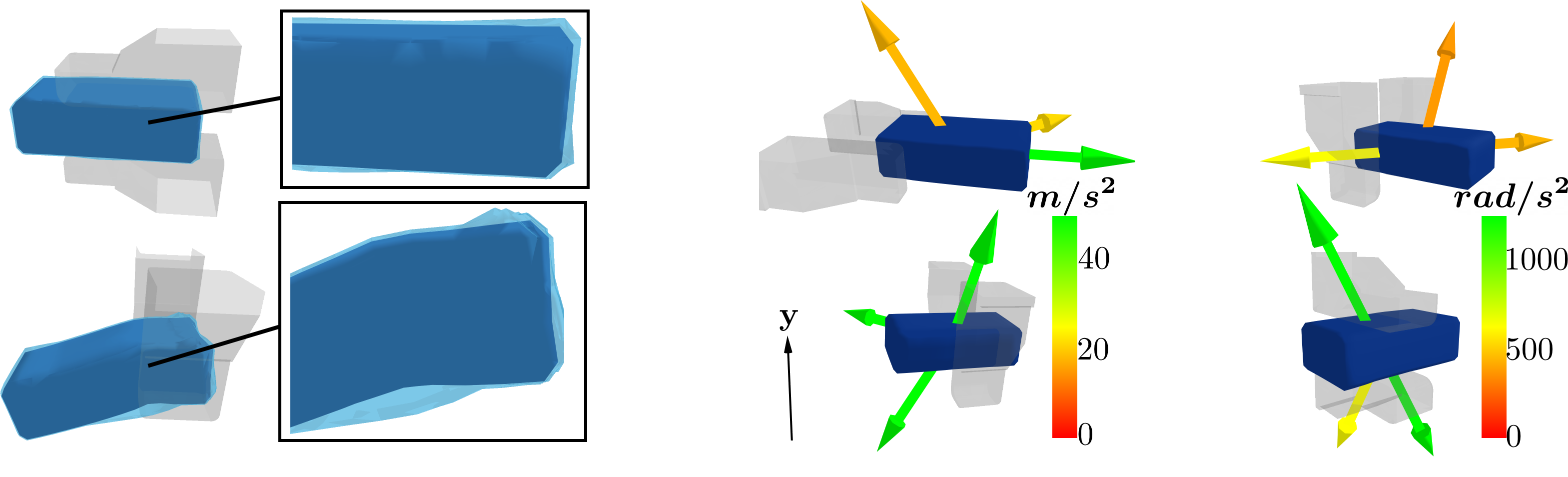}
         \caption{Def. controllability}
         \label{fig:rect_d}
     \end{subfigure}
     \hfill
     \begin{subfigure}[b]{.27\columnwidth}
         \centering
         \includegraphics[scale=0.09, trim={54cm 1.5cm 29cm 0cm},clip]{figures/rectangle/rect_ci3.png}
         \caption{Lin. instability}
         \label{fig:rect_e}
     \end{subfigure}
     \hfill
     \begin{subfigure}[b]{.27\columnwidth}
         \centering
         \includegraphics[scale=0.09, trim={88cm 1.5cm 0cm 0cm},clip]{figures/rectangle/rect_ci3.png}
         \caption{Ang. stability}
         \label{fig:rect_f}
     \end{subfigure}
        \caption{Low (top) and high-valued (bottom) grasps for (a) def. controllability, (b) linear instability, (c) angular instability on a prism ($E{=}2e4$). Controllability plots depict the union of all configurations under reorientation. Instability plots depict the vectors corresponding to the max, min, and median accelerations at failure.}
        \label{fig:rect_ci}
\vspace{-16pt}
\end{figure}

\begin{figure}[t]
     \centering
     \begin{subfigure}[b]{.33\columnwidth}
         \centering
         \includegraphics[scale=0.09, trim={0cm 0cm 62cm 0cm},clip]{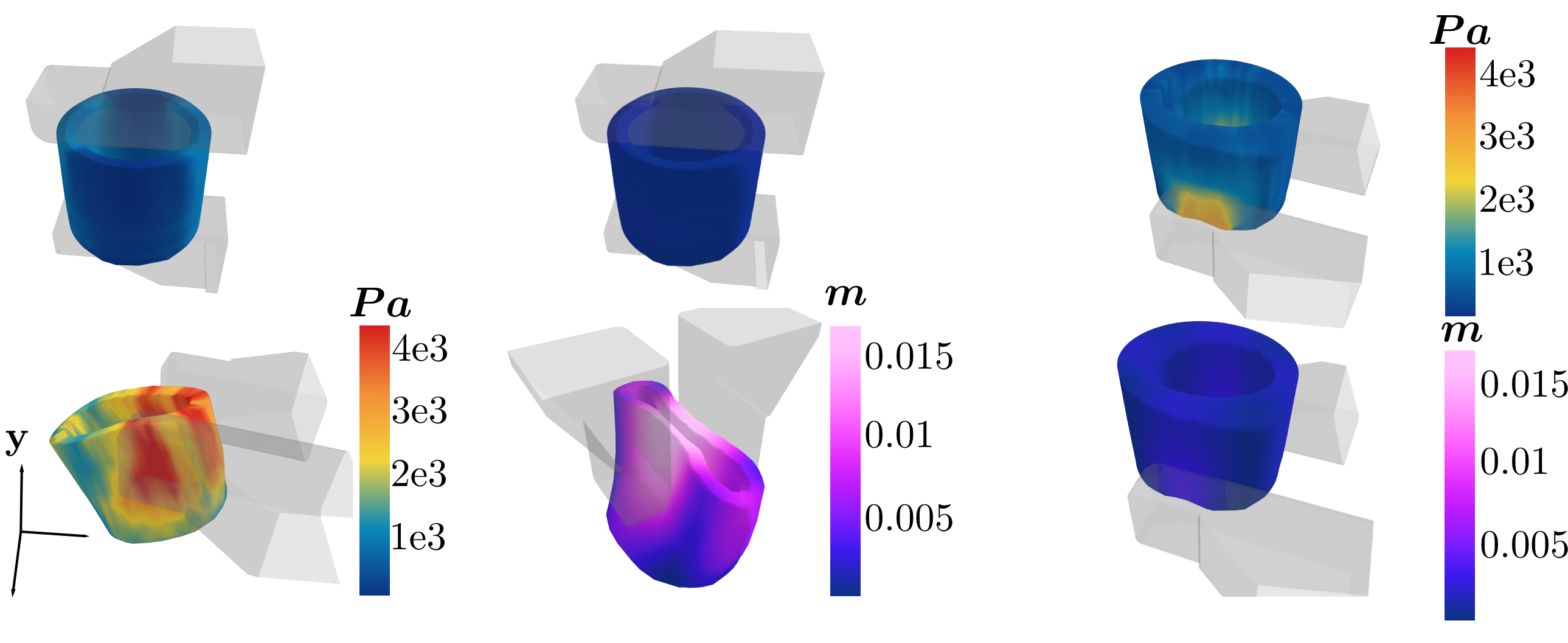}
         \caption{Max stress.}
         \label{fig:cup_a}
     \end{subfigure}
     \hfill
     \begin{subfigure}[b]{.33\columnwidth}
         \centering
         \includegraphics[scale=0.09, trim={32cm 0cm 34cm 0cm},clip]{figures/cup/cup_a.png}
         \caption{Max def.}
         \label{fig:cup_b}
     \end{subfigure}
     \hfill
     \begin{subfigure}[b]{.30\columnwidth}
         \centering
         \includegraphics[scale=0.09, trim={64cm 0cm 0cm 0cm},clip]{figures/cup/cup_a.png}
         \caption{}
         \label{fig:cup_c}
     \end{subfigure}
        \caption{(ab) Low (top) and high-valued (bottom) cup grasps for stress and deformation metrics ($E{=}2e4$); (c) A grasp contacting just the base of the cup can induce high max stress but low max deformation.}
        \label{fig:cup_sd}
\vspace{-18pt}
\end{figure}

For the cup, grasps with low values for maximum stress and deformation squeeze along its height, contacting its base and covering its opening (Fig.~\ref{fig:cup_sd}). In this configuration, the compressive stiffness of the cup is highest, reducing deformation and stress for a given applied force. Grasps with the highest maximum stress and deformation squeeze the sides of the cup near the lip. In this configuration, the cup has the lowest geometric stiffness, increasing deformation and stress. However, additional cases reveal further nuances of grasping cups. In particular, grasps that squeeze the cup on opposite sides of the base induce high stress due to small contact area, but low deformation due to high geometric stiffness (Fig.~\ref{fig:cup_sd}c).

\begin{figure}
     \centering
     \begin{subfigure}[b]{.33\columnwidth}
         \centering
         \includegraphics[scale=0.09, trim={0cm 0cm 58cm 0cm},clip]{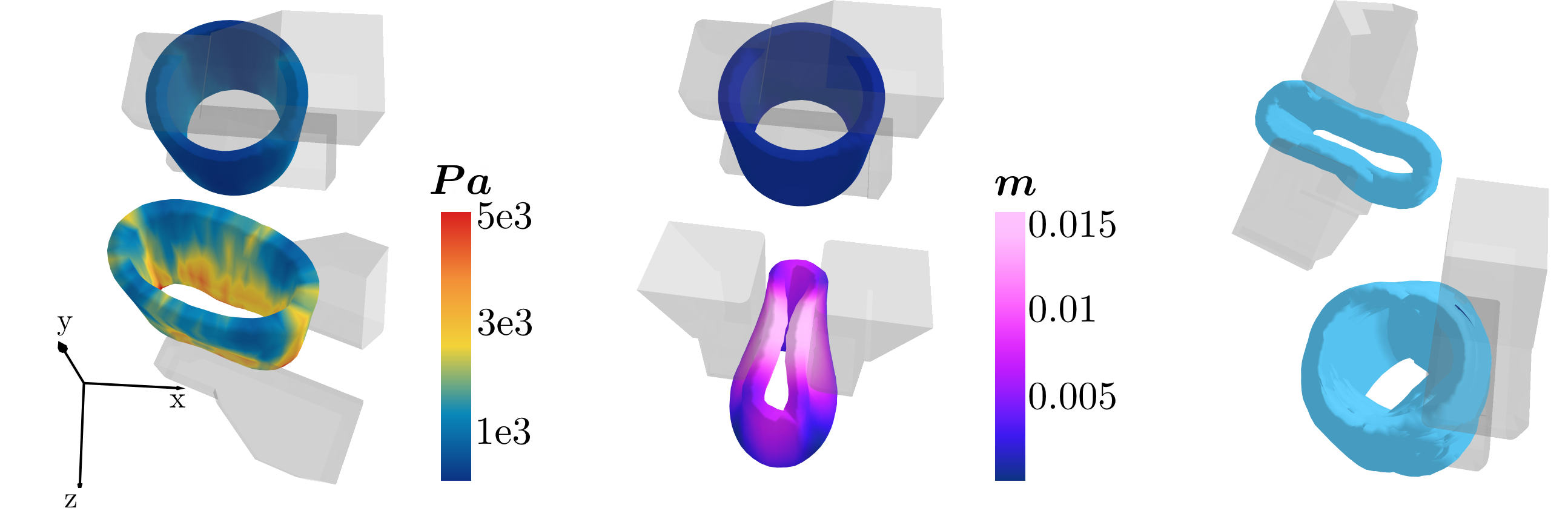}
         \caption{Max stress.}
         \label{fig:ring_a}
     \end{subfigure}
     \hfill
     \begin{subfigure}[b]{.33\columnwidth}
         \centering
         \includegraphics[scale=0.09, trim={36cm 0cm 25cm 0cm},clip]{figures/ring/ring_grasps.png}
         \caption{Max def.}
         \label{fig:ring_b}
     \end{subfigure}
     \hfill
     \begin{subfigure}[b]{.30\columnwidth}
         \centering
         \includegraphics[scale=0.09, trim={69cm 0cm 0cm 0cm},clip]{figures/ring/ring_grasps.png}
         \caption{Def. controllability.}
         \label{fig:ring_c}
     \end{subfigure}
        \caption{Low (top) and high-valued (bottom) ring grasps ($E{=}2e4$).}
        \label{fig:ring_grasps}
        \vspace{-15pt}
\end{figure}

\begin{figure}
     \centering
     \begin{subfigure}[b]{.45\columnwidth}
         \centering
         \includegraphics[scale=0.085, trim={0cm 0cm 60.5cm 0cm},clip]{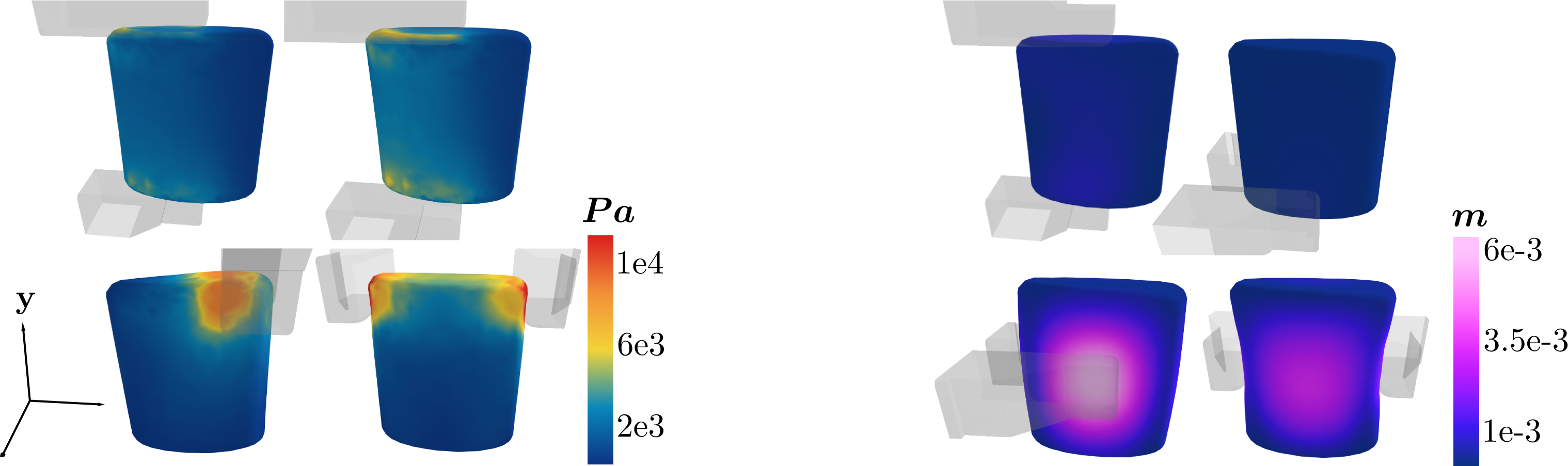}
         \caption{Max stress.}
         \label{fig:flask_a}
     \end{subfigure}
     \hfill
     \begin{subfigure}[b]{.45\columnwidth}
         \centering
         \includegraphics[scale=0.085, trim={63cm 0cm 0cm 0cm},clip]{figures/flask/flask_compact.png}
         \caption{Max deformation.}
         \label{fig:flask_b}
     \end{subfigure}
        \caption{Low (top) and high-valued (bottom) flask grasps ($E{=}2e5$).}
        \label{fig:flask_grasps}
\vspace{-12pt}
\end{figure}

For the ring and flask, grasps with the low values for maximum stress and deformation are top-bottom grasps; as with the cup, in this configuration, the compressive stiffness of these objects is highest, reducing these metrics (Fig.~\ref{fig:ring_grasps}ab and Fig.~\ref{fig:flask_grasps}a). On the other hand, grasps with high values for maximum stress and deformation are side grasps. In this configuration, the ring has the lowest geometric stiffness, allowing buckling and collapse, increasing both bending/compressive stress and deformation. For the flask, the high-stress grasps and high-deformation side grasps occur at slightly different locations. The high-stress grasps squeeze regions with high curvature and low contact area, increasing compressive stress (Fig.~\ref{fig:flask_grasps}a). As with the cup and ring, the high-deformation grasps squeeze in configurations with the lowest geometric stiffness (i.e., perpendicular to the flask face) (Fig.~\ref{fig:flask_grasps}b). 

\begin{figure}
     \centering
     \begin{subfigure}[b]{.16\columnwidth}
         \centering
         \includegraphics[scale=0.088, trim={0cm 0cm 161cm 0cm},clip]{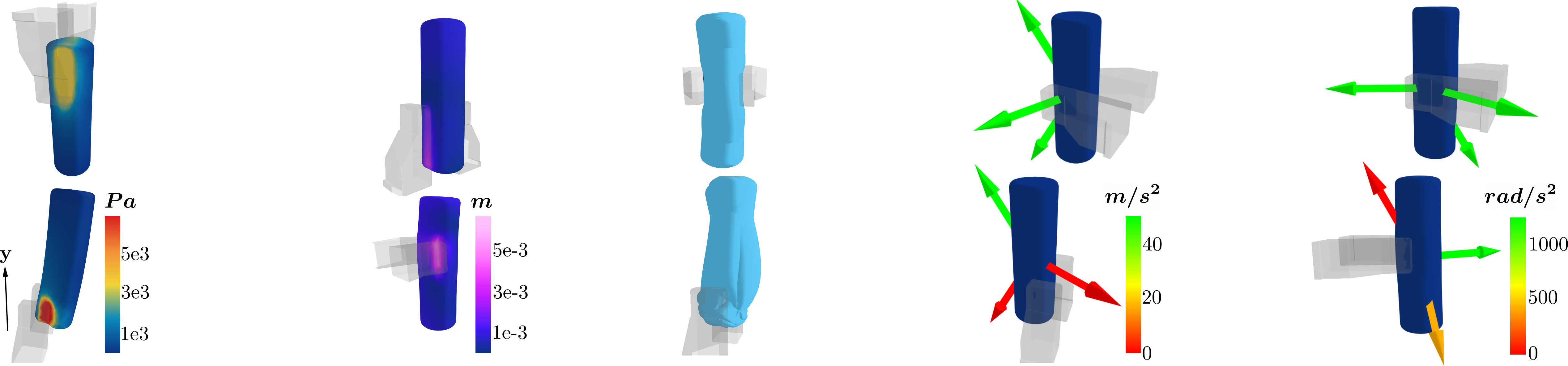}
         \caption{}
     \end{subfigure}
     \hfill
     \begin{subfigure}[b]{.16\columnwidth}
         \centering
         \includegraphics[scale=0.088, trim={42cm 0cm 118cm 0cm},clip]{figures/cylinder/cylinder_all.png}
         \caption{}
     \end{subfigure}
     \hfill
     \begin{subfigure}[b]{.11\columnwidth}
         \centering
         \includegraphics[scale=0.088, trim={77cm 0cm 91cm 0cm},clip]{figures/cylinder/cylinder_all.png}
         \caption{}
     \end{subfigure}
     \hfill
     \begin{subfigure}[b]{.23\columnwidth}
         \centering
         \includegraphics[scale=0.088, trim={111cm 0cm 46cm 0cm},clip]{figures/cylinder/cylinder_all.png}
         \caption{}
     \end{subfigure}
     \begin{subfigure}[b]{.25\columnwidth}
         \centering
         \includegraphics[scale=0.088, trim={150cm 0cm 0cm 0cm},clip]{figures/cylinder/cylinder_all.png}
         \caption{}
     \end{subfigure}
        \caption{Low (top) and high-valued (bottom) grasps for (a) stress, (b) def., (c) def. controllability, (d) lin. instability, (e) ang. instability on a cylinder ($E{=}2e4$).}
        \label{fig:cylinder_grasps}
\vspace{-20pt}
\end{figure}

For the cylinder, grasps with low stress and deformation squeeze along the long axis. In this configuration, the gripper achieves the highest contact area and least thickness between its fingers, lowering compressive stress and deformation (Fig.~\ref{fig:cylinder_grasps}a-b). On the other hand, grasps with high stress and deformation pinch the ends or middle. The gripper has minimal contact area, increasing compressive stress, and bending deformations are minimally restricted, increasing deformation.

\smallskip 

\noindent\textbf{Controllability and Instability Metrics.} We now examine the deformation controllability and linear/angular instability results. For the prism, spheroid, and cylinder, grasps with the highest values for all three of these metrics tend to squeeze the ends of the object. (Fig.~\ref{fig:rect_ci}a-c, Fig.~\ref{fig:spheroid_grasps}c, and Fig.~\ref{fig:cylinder_grasps}c). Here, the gripper allows cantilever bending with a maximally long moment arm, maximizing deformation controllability (i.e., deformation under reorientation). In addition, the gripper is maximally close to losing contact with the object under displacement perturbation, increasing linear and angular instability. For the cylinder, these grasps also coincide with those for the highest values for maximum stress and deformation.

\begin{figure}
     \centering
     \begin{subfigure}[b]{.3\columnwidth}
         \centering
         \includegraphics[scale=0.09, trim={3cm 0cm 68cm 0cm},clip]{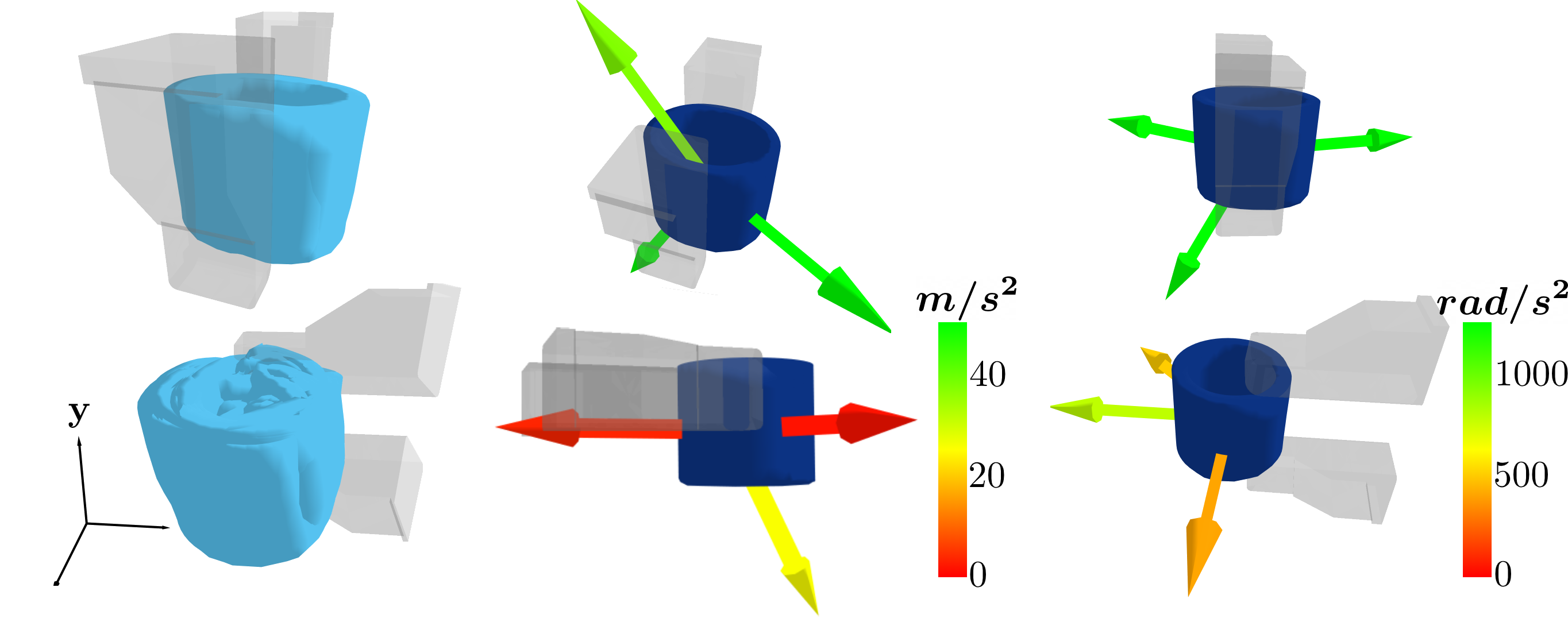}
         \caption{Def. controllability.}
         \label{fig:cup_d}
     \end{subfigure}
     \hfill
     \begin{subfigure}[b]{.33\columnwidth}
         \centering
         \includegraphics[scale=0.09, trim={30cm 0cm 34cm 0cm},clip]{figures/cup/cup_b_updated.png}
         \caption{Lin. instability.}
         \label{fig:cup_e}
     \end{subfigure}
     \hfill
     \begin{subfigure}[b]{.33\columnwidth}
         \centering
         \includegraphics[scale=0.09, trim={64cm 0cm 0cm 0cm},clip]{figures/cup/cup_b_updated.png}
         \caption{Ang. instability.}
         \label{fig:cup_f}
     \end{subfigure}
        \caption{Low (top) and high-valued (bottom) cup grasps for controllability and instability metrics ($E{=}2e4$).}
        \label{fig:cup_ci}
\end{figure}

For the cup, grasps with low values for deformation controllability and linear and angular instability tend to make substantial contact with the object, particularly its base (Fig.~\ref{fig:cup_ci} top). In this configuration, the gripper constrains the degrees of freedom of the object, reducing controllability; furthermore, the gripper contacts locally stiff regions, enabling fast response to dynamic perturbations and reducing instability. 

For the ring and flask, grasps with the highest values for deformation controllability and linear and angular instability have small amounts of material between the gripper fingers. In this configuration, the gripper exposes more degrees of freedom of the object, increasing deformation under reorientation and sensitivity to displacement perturbation. For example, for the flask, high-valued grasps squeeze along the shortest axis on areas with high curvature (not depicted). Conversely, grasps with the lowest deformation controllability and linear and angular instability have large amounts of material between the gripper fingers. In this configuration, the gripper restricts more degrees of freedom. For example, for the ring, low-valued grasps squeeze along the side of the ring towards the ring's central axis, leaving only a small surface uncontacted.

\subsection{Feature Importance and Predictive Power}
For each primitive with a specified elastic modulus ($E$), we evaluate 7 grasp features and 7 performance metrics for all grasps. We then examine the power of the features to predict the metrics. For each continuous-valued performance metric (i.e., all metrics except pickup success), we build a random forest classifier. The classifier takes as input the grasp features, and outputs whether the corresponding grasp belongs to the top or bottom 30th percentile of all grasps, ranked by their metric values; binary classification is performed because separating the two extremes reveals highly interpretable physical trends.

We use random forests for their ability to handle relatively small training sets, as well as their prior successes in predicting grasp outcomes~\cite{Rubert2017IROS}. The predictive power of each grasp feature is then quantified as its Gini impurity-based importance. For each performance metric, we initially trained separate classifiers for each $E$; however, the relative feature importances were qualitatively similar over all moduli. We thus combine grasp samples across all $E$ to train one model per metric and simply add $E$ as a feature.

\begin{figure*}
\centering
\includegraphics[scale=0.14, trim={0cm 0cm 0cm 0cm},clip]{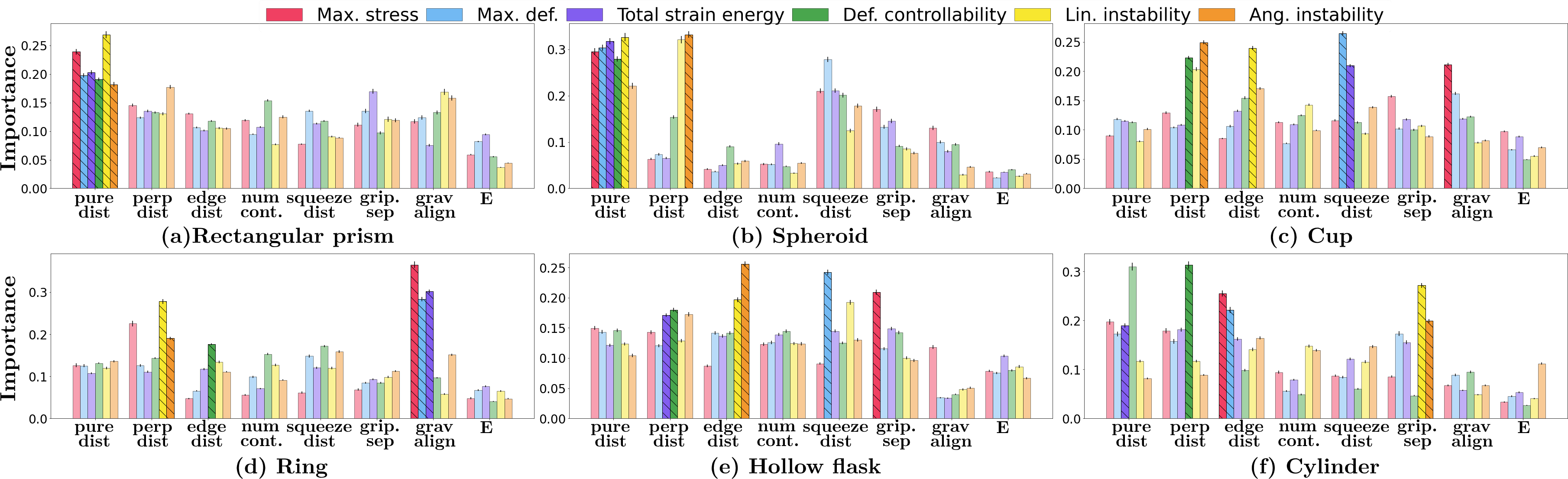}
\caption{Feature importance across performance metrics for each primitive. Hatched bars denote the most important feature for each metric; error bars (small) denote the standard error. }
\label{fig:imps}
\vspace{-18pt}
\end{figure*}

Fig.~\ref{fig:imps} depicts the feature importance of the resulting classification models; critically, the importance of each feature varies across the primitives and metrics. For the prism and spheroid, the $pure\_dist$ feature has the highest importance for all metrics. This aligns with our observation that for the grasps with the highest values for each performance metric, the gripper tends to squeeze the ends of the object. However, $perp\_dist$ is also an important feature for angular instability on the prism, and is the most important feature for this metric on the spheroid. Rotational perturbations are centered between the fingertips, and $perp\_dist$ directly captures the length of the moment arm. 

Between the prism and spheroid, a key difference is that the pickup success rate for the spheroid drops dramatically from $90\%$ at $E{=}2e6$ to $12\%$ at $E=2e9$. The ring, cup, and cylinder also experience a drop in pickup success at high $E$. When these rounded objects are stiff, the gripper becomes unable to induce deformation and generate sufficient contact area upon squeezing; thus, grasps such as those in the bottom row of Fig.~\ref{fig:spheroid_grasps}c easily drive the object out of the gripper.

For the cup, $grav\_align$ has the highest importance for predicting maximum stress. This aligns with our observation that the grasps with the lowest maximum stress are all top-bottom grasps. In addition, $squeeze\_dist$ is most important for predicting maximum deformation. As the gripper squeezes the object, this feature directly characterizes the stiffness of the material between the fingers. In contrast to the prism and spheroid, $perp\_dist$ has substantially higher importance than $pure\_dist$ for predicting deformation controllability and instability. When the gripper squeezes the cup, $perp\_dist$ stably measure of the distance between the gripper and the opening of the cup, which is critical for controllability and stability. However, $pure\_dist$ changes substantially, regardless of how far the gripper is from the opening.

For the ring, $grav\_align$ feature has the highest importance for predicting all deformation-related metrics.
Although the deformations induced by side grasps on a cup vary based on height (i.e., side grasps at the base induce less deformation than at the opening), all side grasps on the ring induce high deformation. 
Thus, the grasps with the lowest and highest values for deformation-related metrics are best distinguished by whether they are top-bottom or side grasps, and $grav\_align$ precisely captures this. In addition, $edge\_dist$ and $num\_contacts$ have high importance for predicting deformation controllability. This aligns with our observation that the amount of material squeezed by the gripper greatly influences controllability. This quantity depends on both $edge\_dist$ (distance between contact patch and gripper finger edge) and $num\_contacts$ (contact area). As with the cup, $perp\_dist$ is most important for predicting linear and angular instability.

For the flask, $gripper\_sep$ has the highest importance for predicting maximum stress. As with the ring, the grasps with the lowest and highest stress are distinguished by whether they are top-bottom or side grasps. In addition, as with the cup, $squeeze\_dist$ is most important for predicting deformation
Finally, $edge\_dist$ is dominant for predicting linear and angular instability. This feature quantifies the physical displacement required for the object to lose contact with the gripper.

For the cylinder, $edge\_dist$ has the highest importance for predicting maximum stress and deformation. On this geometry, this feature quantifies how strongly the grasp pinches the ends or middle of the object. In addition, $pure\_dist$ and $perp\_dist$ are highly correlated ($R$=0.99); all candidate grasps squeeze perpendicular to the long axis of the cylinder, resulting in negligible difference between the two features. These features are most important for predicting deformation controllability. They quantify the distance of the gripper to the ends of the object, determining the moment arm for bending.

\subsection{Fundamental Trends}
Based on our analysis of object primitives, we identify several fundamental trends. Stress is high for low gripper-object contact area or when inducing object buckling; deformation is high for low object geometric stiffness.
While stress and deformation are related, high-stress and high-deformation grasps can be dissimilar (e.g. Fig.~\ref{fig:cup_sd}\subref{fig:cup_c}). Instability and deformation controllability are maximized when the gripper contacts one end of the object, as the object can be displaced or exhibit gravity-induced deformation. Our importance analyses also consistently identify features that predict these metrics and agree with physical reasoning. Although elastic modulus impacts the set of successful grasps (e.g., failure for stiff versions of rounded objects), it does not substantially impact how grasps rank according to the metrics.


\section{Sim-to-Real Accuracy}\label{app:sim2real}
\begin{figure}
\centering
\includegraphics[scale=0.12, trim={0cm 0cm 0cm 0cm},clip]{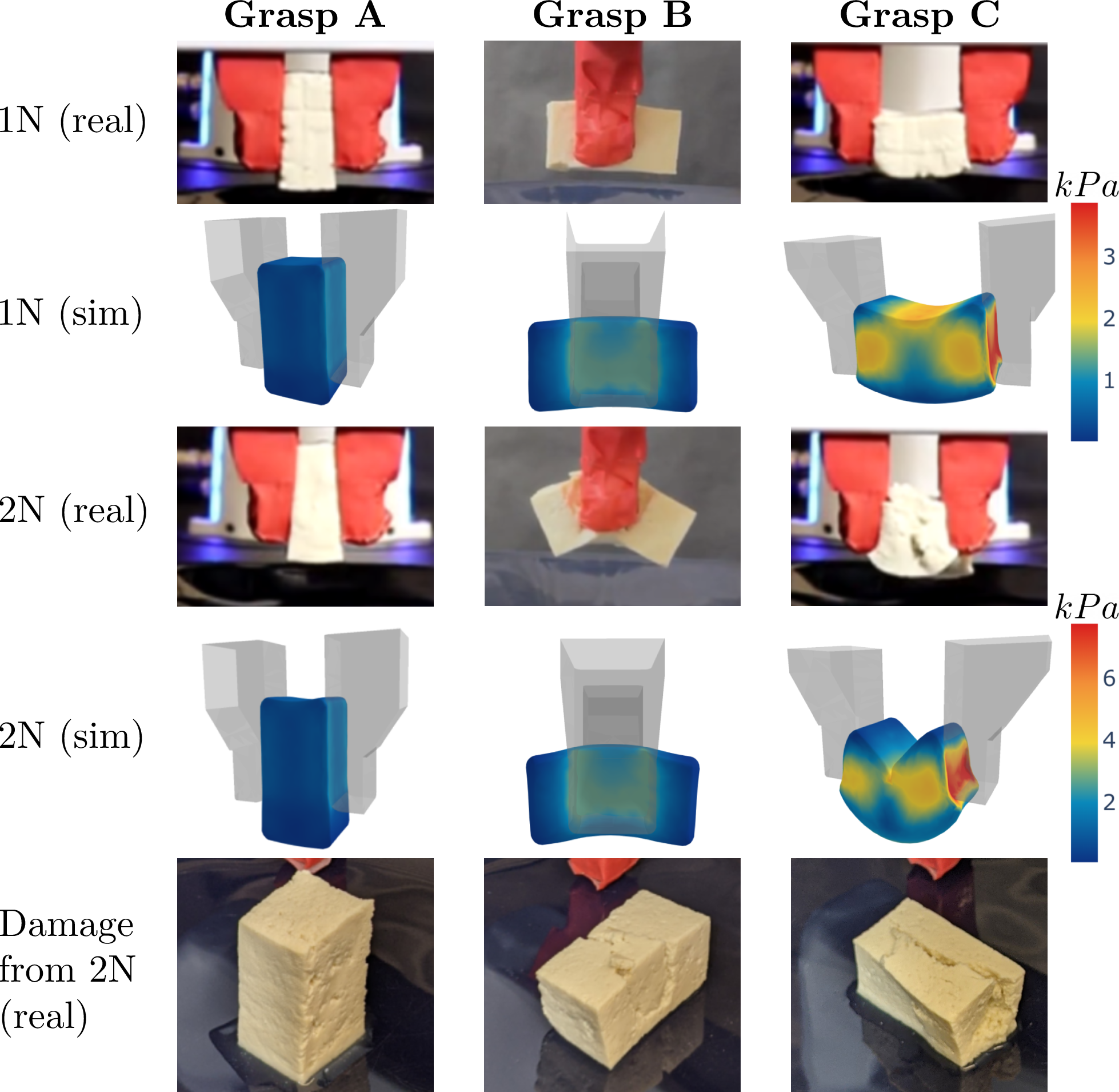}
\caption{Three grasps tested on blocks of tofu (1 and 2~N of squeezing force) show similar outcomes in simulation and the real world.}
\label{fig:tofu_grasps}
\vspace{-6pt}
\end{figure}

If our simulations prove to be accurate to the real-world, we can proceed with confidence in applying simulation-generated grasp strategies to reality. We test 3 grasps on real blocks of tofu under 1 and 2~N of applied force~(see~Fig.~\ref{fig:tofu_grasps}). Simulated and real-world deformations exhibit strong similarities, and grasps achieve anticipated performance (e.g., grasps~A and~C,  respectively, minimize and maximize sagging under the 2 force conditions). In the real world, permanent damage on the tofu occurs under 2~N of applied force, with fracture occurring in grasps~B and~C. Although fracture cannot be simulated with FEM, simulated stresses for these grasps lie within the literature-reported range of breaking stress for tofu~\cite{Toda2003SeedPC}; furthermore, fracture lines on the real tofu coincide with regions of high simulated stress. 

\begin{figure}
\centering
\includegraphics[scale=0.3, trim={0cm 0cm 0cm 0cm},clip]{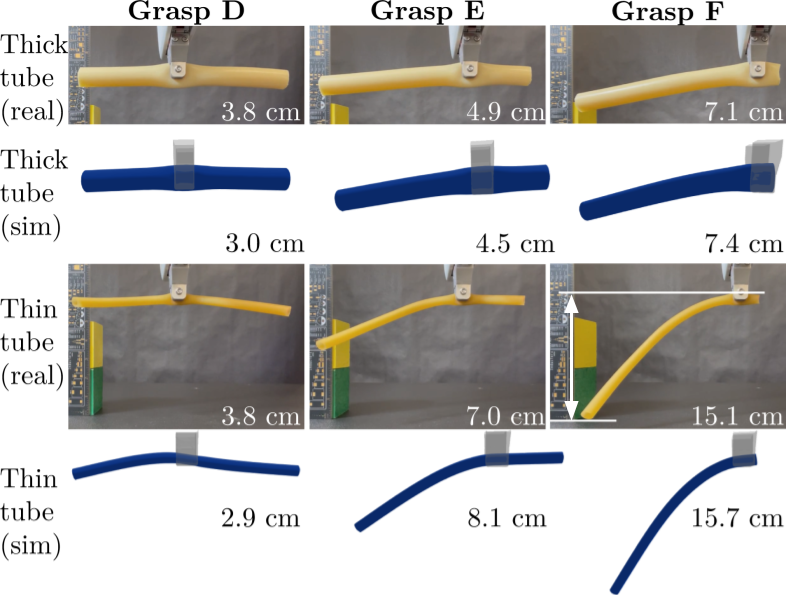}
\caption{Three grasps tested on 2 real and simulated latex tubes under 15 N of gripper force. The vertical distance between the highest and lowest points of the tube are shown alongside each image. (See Grasp F, bottom row for an example.) Localized deformation due to compression at the grippers is also replicated in simulation.}
\label{fig:hollow_tube_grasps}
\end{figure}

Next, we perform 3 grasps on 2 latex tubes of different geometry (Fig.~\ref{fig:hollow_tube_grasps}). Again, simulated and real-world deformations are highly similar (including indentations and bulges localized to the contact locations); moreover, the vertical distance between the highest and lowest points of the tubes closely match. In simulation, deformation controllability is predicted to be higher in grasp F than grasp D (Fig.~\ref{fig:cylinder_grasps}c). Upon reorienting the rubber tubes in the real world, grasp F indeed enables a sequence of higher maximum deformations than grasp D, which constrains the tube to closely follow the motion of the gripper (thus, sweeping out a larger angle).


\begin{figure}
\centering
\includegraphics[scale=0.4, trim={0cm 0cm 0cm 0cm},clip]{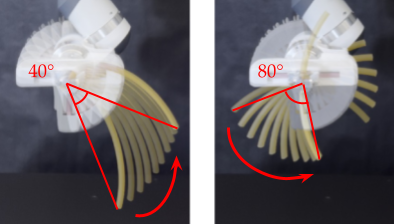}
\caption{Two tube grasps under a 90-degree rotation of the gripper: (left) an end grasp, and (right) a middle grasp. The angle swept out by the tip of the tube is marked on both images; the arrow denotes the counterclockwise direction of rotation.}
\label{fig:tube_reorient}
\end{figure}

\end{appendices}
\end{document}